\documentclass[10pt]{article}

\usepackage[a4paper, margin=0.9in]{geometry}
\usepackage{graphicx}
\usepackage{hyperref}
\usepackage[nolist]{acronym}
\newacro{kNN}{k-Nearest Neighbours}
\newacro{POS}{part-of-speech}
\newacro{NER}{named entity recognition}
\newacro{NLP}{natural language processing}
\newacro{tfidf}[tf-idf]{term frequency-inverse document frequency}

\newacro{20news}[20News]{\textit{20 News Group}}
\newacro{moviereviews}[Reviews]{\textit{Movie Reviews}}
\newacro{reuters8}[Reuters-8]{\textit{Reuters-8}}
\usepackage{amsmath,amssymb}
\usepackage{calrsfs}
\DeclareMathAlphabet{\pazocal}{OMS}{zplm}{m}{n} 
\usepackage{cleveref}

\usepackage{natbib}
\setcitestyle{authoryear,open={(},close={)}}

\title{Domain-adaptation of spherical embeddings}
\author{Mihalis Gongolidis\textsuperscript{1}, Jeremy Minton\textsuperscript{1}, Ronin Wu\textsuperscript{1},\\ Valentin Stauber\textsuperscript{1}, Jason Hoelscher-Obermaier\textsuperscript{1} and Viktor Botev\textsuperscript{1}\\
\small{\textsuperscript{1}Iris AI, Bekkestua, Norway}}
\date{}

\begin{document}


\newcommand{\pos}[0]{\mathbf{u}}  
\renewcommand{\neg}[0]{{\mathbf{u}^\prime}}  
\newcommand{\doc}[0]{\mathbf{d}}  
\newcommand{\ctx}[0]{\mathbf{v}}  
\newcommand{\loss}{\pazocal{L}}  

\newcommand{\dcos}{d_\mathrm{cos}}  

\renewcommand{\cos}[1]{\mathrm{cos} \left( {#1} \right)}

\newcommand{\identity}{\ensuremath{\mathbb{I}}}
\newcommand{\Tr}{\mathrm{Tr}}

\maketitle

Domain adaptation of embedding models, updating a generic embedding to the language of a specific domain, is a proven technique for domains that have insufficient data to train an effective model from scratch.
Chemistry publications is one such domain, where scientific jargon and overloaded terminology inhibit the performance of a general language model.

The recent spherical embedding model (JoSE) proposed in \citet{meng_spherical_2019} jointly learns word and document embeddings during training on the multi-dimensional unit sphere, which performs well for document classification and word correlation tasks. But, we show a non-convergence caused by global rotations during its training prevents it from domain adaptation.

In this work, we develop methods to counter the global rotation of the embedding space and propose strategies to update words and documents during domain specific training.
Two new document classification data-sets are collated from general and chemistry scientific journals to compare the proposed update training strategies with benchmark models. We show that our strategies are able to reduce the performance cost of domain adaptation to a level similar to Word2Vec.

\section*{Embedding space tumbling}
    A global rotation of the embedding space was observed during training and can be confirmed by the magnitude of updates between epochs, which converge to a non-zero value.
    While the impact this has on the quality of an embedding is empirically bounded by task performance, it is detrimental to domain adaptation.
    Any document or word vector that is not updated will become misaligned by the global rotation, which is much larger than the typical adaptation to specific domains.

    Investigation revealed a discrepancy between the method described in \citet{meng_spherical_2019} and the provided code. Specifically, the adjustment factor, $\dcos$, in the gradient retraction onto the spherical space differs. 
    It can be shown that the factors described in the paper reduce to $0$ or $1$. The $\dcos$ equations implemented in the supplied code do not, but we could not find justification for those values. It was decided to drop the $\dcos$ factor entirely and this halved the update magnitude the model converge to.

    This change is a significant improvement, but does not entirely eliminate the global rotation so a post-processing solution is introduced.
    A counter-rotation of the embedding space to minimize the sum of angles between its vectors and reference embedding can be determined with a singular value decomposition.
    This is a feasible approach to eliminate global rotation relative to a reference embedding, chosen to be after a given number of epochs.
    
\section*{Document embedding strategies}

    The propos    A Word2Vec\citep{mikolov_efficient_2013} model is used as a benchmark, where word vectors are averaged in place of the document vector.

ed loss function in the original spherical embeddings model as discussed in \citet{meng_spherical_2019} does not allow continuation of training for document vectors. 
    To enable it when adapting the generic embeddings model with a domain-specific corpus, two approaches are considered:
    
    \begin{enumerate}
        \item In the first approach, we compress each document to $n$ keywords, selected as the most cosine similar words to the document vector.
        This compressed representation is added as cosine similarities to the loss function so the generic document vectors are updated during domain adaptation. This makes the loss function
        \begin{equation}
            L_{u \in G \cap D} = L_{Global} + L_{Local} +\alpha \sum_{d \in P_{u}}\mathrm{max}(0,m-cos(d,u)+cos(d,n))
        \label{eq:adapted-loss}
        \end{equation}
        where the nomenclature follows Equation 3 of \citet{meng_spherical_2019} as well as the newly introduced terms $P_{u}$ representing the set of document vectors a word vector, $u$, is represented with and a weighting parameter $\alpha$.
        
        An advantage of this method, is that keywords that only appear in the original corpus continue to be updated.
        
        \item The second approach constructs the document vectors from an average of word vectors, replacing, $d_i$, in the loss function with $\frac{1}{|D_i|} \sum_{n \in D_i} u_n$. This is more efficient than the previous approach because the dynamically constructed document vectors are efficiently computed and do not need to be persisted and updated.
    \end{enumerate}

    We will compare these approaches with document classification, using a \ac{kNN} classifier ontop of the document vectors from each model.
 
    Two new data-sets, from https://core.ac.uk/, were collated to test these embedding models:
    \begin{description}
        \item[BioChem] Approximately 280 thousand biochemical research abstracts across 18 biochemistry subtopics based on the author's keywords. The subtopics are evenly distributed in both train and update corpora.
        \item[Core-general] Approximately 210 thousand research abstracts from 14 topics and three chemistry related topics. The update corpus is composed of the three chemistry topics and the train corpus is composed of the rest.
    \end{description}
    These were preprocessed with tokenization, lowercasing and lemmatization.
    In addition, three open-source data-sets were used: \ac{20news}\citep{news20}, \ac{moviereviews}\citep{moviereviews} and \ac{reuters8}\citep{meng_unsupervised_2020}.
    Each dataset is divided into training, update and test splits.
    As a reference, the benchmark models are trained on the combined training and update splits.
    Then, to simulate domain adaptation, the benchmark and newly proposed approaches were trained on the training split, before separately being updated on the update split.
    
    A Word2Vec\citep{mikolov_efficient_2013} model is used as a benchmark, where word vectors are averaged in place of the document vector.
    
    \begin{table}[h]
    \centering
            \begin{tabular}{l|ccccc}
            Training Approach        & \acs{20news}   & \acs{reuters8}   & \acs{moviereviews}   & BioChem        & Core-general   \\
            \hline
            Word2vec                 & \textbf{0.69}          & \textbf{0.90}  &  0.75          & \textbf{0.62}  & 0.76  \\
            JoSE                     & 0.68                   & 0.88           &  \textbf{0.77} & 0.60           & \textbf{0.78}  \\
            \hline
            Retraining Approach     & \acs{20news}    & \acs{reuters8}   & \acs{moviereviews}  & BioChem        & Core-general   \\
            \hline
            Word2vec                & \textbf{0.70}   & 0.89             & \textbf{0.77}       & \textbf{0.62}  & 0.71           \\
            JoSE                    & 0.62            & 0.78             & 0.72                & 0.54           & 0.59           \\
            JoSE (Compression)      & 0.66            & 0.86             & 0.71                & 0.59           & 0.73           \\
            JoSE (Construction)     & 0.65            & \textbf{0.89}    & 0.74                & 0.60           & \textbf{0.76}  \\
            \hline
        \end{tabular}
        \caption{Macro F1 scores for initially training on the training set and applying subsequent training on the test set for the different update approaches (top) and different construction approaches (bottom).}
        \label{tab:results}
    \end{table}
    
    \Cref{tab:results} shows both our proposed approaches improve on naive retraining, reducing the perofrmance cost of domain-adaptation from 12\% to 2\% for the construction approach. This is similar to the performance drop of Word2Vec, which is around 0.7\%.
    
    Of the two proposed approaches, the construction approach performed better, outperforming the other JoSE models on four of the five datasets. This is particularly attractive given it is computationally faster than computing the word-document linkages.
    
\section*{Conclusion}
    In this work, we develop the JoSE model presented in \citet{meng_spherical_2019} to accept update training steps for applications such as domain adaptation. This required the elimination of spinning of the embedding space during training, which was achieved by correcting a likely error in the original model and applying a post-processing counter-rotation.
    Further, two approaches were proposed to account for the inability to update document vectors during update training steps: adding compressed document representations to the loss function and constructing document-vectors as averaged word-vectors.
    Two novel data-sets of scientific publications for document classification, specifically focusing on chemistry were collated for testing against benchmark models.
    These approaches improved the JoSE model's document classification performance after update training. The performance against a Word2Vec benchmark in this context is now consistent with the performance presented in \citet{meng_spherical_2019}. The results show that our approaches are able to reduce the cost of domain adaptation to a level similar to Word2Vec and hence this modified JoSE model is now competitive with Word2Vec in domain adaptation tasks.

\bibliography{bibliography}

\end{document}